\def\year{2019}\relax
\documentclass[letterpaper]{article} 
\usepackage{aaai19}  
\usepackage{times}  
\usepackage{helvet}  
\usepackage{courier}  
\usepackage{url}  
\usepackage{graphicx}  

\usepackage{booktabs} 
\usepackage{fancyhdr}
\usepackage{amssymb}
\usepackage{booktabs}
\usepackage{enumitem}
\usepackage[flushleft]{threeparttable}
\usepackage{epsfig}
\usepackage{epstopdf}
\usepackage{graphicx}
\usepackage{ifpdf}
\ifpdf
  \DeclareGraphicsExtensions{.pdf,.png,.jpg}
\else
  \DeclareGraphicsExtensions{.eps}
\fi
\usepackage{mathrsfs}
\usepackage{amsmath}
\usepackage{algorithmic}
\usepackage{array}
\usepackage{marvosym}
\usepackage{textcomp}
\usepackage{times}
\usepackage{soul}
\usepackage[utf8]{inputenc}
\usepackage{array}
\usepackage{bbm}

\author{Yitian Yuan$^1$, Tao Mei$^2$, Wenwu Zhu$^{1,3}$\\
$^1$ Tsinghua-Berkeley Shenzhen Institute, Tsinghua University, China \\
$^2$ JD AI Research, China\\
$^3$ Department of Computer Science and Technology, Tsinghua University, China\\
yyt18@mails.tsinghua.edu.cn, tmei@jd.com, wwzhu@tsinghua.edu.cn\\
}

\begin{document}
%
\title{To Find Where You Talk: Temporal Sentence Localization in Video \\ with Attention Based Location Regression}


\maketitle

\begin{abstract}
We have witnessed the tremendous growth of videos over the Internet, where most of these videos are typically paired with abundant sentence descriptions, such as video titles, captions and comments. Therefore, it has been increasingly crucial to associate specific video segments with the corresponding informative text descriptions, for a deeper understanding of video content. This motivates us to explore an overlooked problem in the research community --- temporal sentence localization in video, which aims to automatically determine the start and end points of a given sentence within a paired video. For solving this problem, we face three critical challenges: (1) preserving the intrinsic temporal structure and global context of video to locate accurate positions over the entire video sequence; (2) fully exploring the sentence semantics to give clear guidance for localization; (3) ensuring the efficiency of the localization method to adapt to long videos. To address these issues, we propose a novel Attention Based Location Regression (ABLR) approach to localize sentence descriptions in videos in an efficient end-to-end manner. Specifically, to preserve the context information, ABLR first encodes both video and sentence via Bi-directional LSTM networks. Then, a multi-modal co-attention mechanism is presented to generate both video and sentence attentions. The former reflects the global video structure, while the latter highlights the sentence details for temporal localization. Finally, a novel attention based location prediction network is designed to regress the temporal coordinates of sentence from the previous attentions. We evaluate the proposed ABLR approach on two public datasets ActivityNet Captions and TACoS. Experimental results show that ABLR significantly outperforms the existing approaches in both effectiveness and efficiency.

\end{abstract}

\section{Introduction}
\begin{figure}
\centering
\includegraphics[width=3.2in]{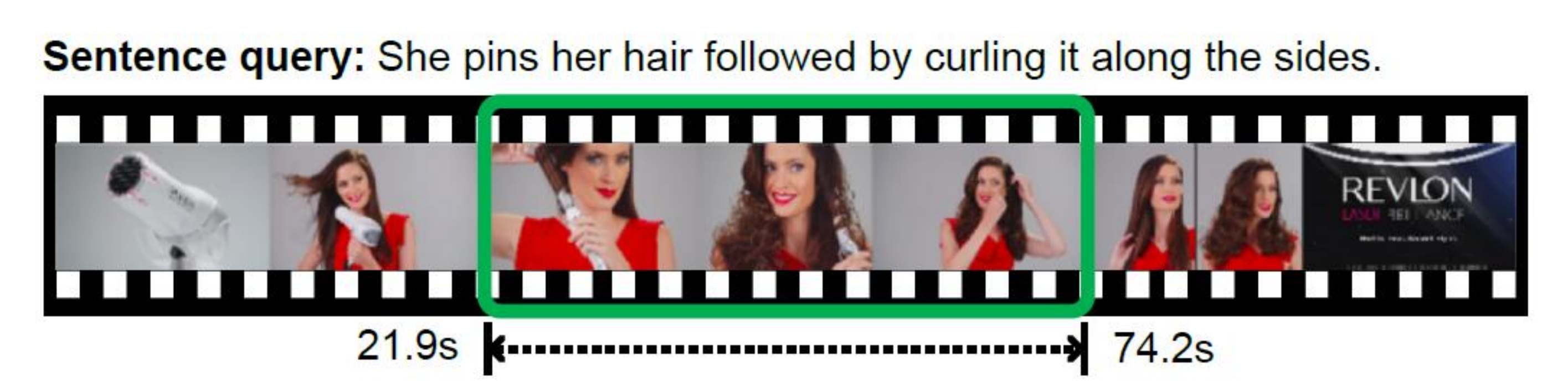}
\caption{ \small Temporal sentence localization in untrimmed video.}
\label{fig:introduction}
\end{figure}

Video has become a new way of communication between Internet users with the proliferation of sensor-rich mobile devices. Moreover, as videos are often accompanied by text descriptions, e.g., titles, captions or comments, it has encouraged the development of advanced techniques for a broad range of video-text understanding applications, such as video captioning \cite{Pan2016Jointly,Duan2018WSDEC}, video generation conditioned on captions \cite{Pan2017To} and video question answering \cite{Xu2017VideoQA}. While promising results have been achieved, one fundamental issue underlying these technologies is overlooked, i.e., the informative video segments should be trimmed and aligned with the relevant textual descriptions. This motivates us to investigate the problem of temporal sentence localization in video. Formally, as shown in Figure \ref{fig:introduction}, given an untrimmed video and a sentence query, the task is to identify the start and end points of the video segment in response to the given sentence query.

To solve the temporal sentence localization in video, people may first consider applying the  typical multi-modal matching architecture \cite{Hendricks2017Localizing,Bojanowski2015Weakly,Gao2017TALL,liu2018attentive}. One can first sample candidate clips by scanning videos with various sliding windows, then compare the sentence query with each of these clips individually in a multi-modal common latent space, and finally choose the highest matched clip as the localization result. While simple and intuitive, this ``scan and localize''  architecture still has certain limitations as follows. First, independently fetching video clips may break the intrinsic temporal structure and global context of videos, making it difficult to holistically predict locations over the entire video sequence. Second, as the whole sentence is represented with a coarse feature vector in the multi-modal matching procedure, some important sentence details for temporal localization are not fully explored and leveraged. Third, densely sampling sliding window is computationally expensive, which limits the efficiency of the localization method in applying to long daily videos in practice.

To address the above limitations, we consider that temporal sentence localization should be strengthened from both video and sentence aspects. From the video aspect, compared with partially watching each candidate clip, it is more natural for people to look through the whole video and then decide which part does the sentence describe. In the latter case, the global video context is intact and the computation overload caused by densely scanning is avoided, resulting in a more comprehensive and efficient video understanding. Therefore, predicting the sentence position upon the overall video sequence from a global view is better than the local matching strategy upon each candidate video clip. From the sentence aspect, as some words or phrases can give clear cues to identify the target video segment, we should pay more attention to these sentence details so as to predict more accurate locations.

Based on the above considerations, we propose an end-to-end Attention Based Location Regression (ABLR) model for temporal sentence localization. The proposed ABLR model goes beyond the existing ``scan and localize'' architecture, and can directly output the temporal coordinates of the localized video segment. Specifically, ABLR first utilizes two Bi-directional LSTM networks to encode video clip and word sequences respectively, where each unit is enriched by the flexible forward and backward context. Based on the encoded features, a multi-modal co-attention mechanism is designed to learn both video and sentence attentions. Video attention incorporates the relative associations between different video parts and the sentence query, and therefore it can reflect the global temporal structure of video. Sentence attention highlights the crucial details in the word sequence, and therefore it can give clear guidance for the following temporal location prediction. Finally, a multilayer attention based location prediction network is proposed to regress the temporal coordinates of the target video segment from the previous global attention outputs. By jointly learning the overall model, our ABLR is able to localize sentence query in video efficiently and effectively.

The main contributions of this paper are summarized as follows:

(1) We address the problem of temporal sentence localization in video by proposing an effective end-to-end Attention Based Location Regression (ABLR) approach. Different from the existing ``scan and localize'' architecture which partially processes each video clip, our ABLR directly predicts the temporal coordinates of sentence queries from a global video view.

(2) We introduce the multi-modal co-attention mechanism for temporal localization task. The multi-modal co-attention mechanism leverages the sentence features to divert the attention to the most indicative video parts, and meanwhile investigates the important sentence details for localization.

(3) We conduct experiments on two public datasets ActivityNet Captions \cite{Krishna2017Dense} and TACoS \cite{Regneri2013Grounding}. The results demonstrate our proposed ABLR model not only achieves superior localization accuracy, but also boosts the localization efficiency compared to the existing approaches.

\section{Related Work}

We briefly group the related works into two main directions: temporal action localization and temporal sentence localization. The former direction aims to solve the problem of recognizing and determining temporal boundaries of action instances in untrimmed videos. Although promising results have been achieved \cite{Shou2016Action,Lin2017Single,Escorcia2016DAPs}, one major limitation is that they are restricted to a predefined list of action categories, which cannot precisely identify the complex scenes and activities in videos. Therefore, some researchers begin to explore the latter direction: temporal sentence localization in video, which is also the main focus of this paper.

Localizing sentences in videos is a challenging task which requires both language and video understanding. Early works mainly constrain to certain visual domains (movie scenes, laboratory or kitchen environment), and often focus on localizing multiple sentences within a single video in chronological order. Inspired by the Hidden Markov Model, Naim and Song \emph{at al.} proposed unsupervised methods to localize natural language instructions to corresponding video segments \cite{Naim2014Unsupervised,song2016unsupervised}. Tapaswi \emph{et al.} computed an alignment between book chapters and movie scenes using matching dialogs and character identities as cues with a graph based algorithm \cite{Tapaswi2015Book2Movie}. Bojanowski \emph{et al.} proposed weakly supervised alignment model under ordering constrain. They cast the alignment between video clips and sentences as a temporal assignment problem, and learned an implicit linear mapping between the vectorial features of the two modalities \cite{Bojanowski2015Weakly}. In contrast to the above approaches, we aims to solve the temporal sentence localization problem for general videos without any domain restrictions. Moreover, we do not rely on the chronological order between different sentence descriptions, i.e., each sentence is independent in the localization procedure.

For localizing independent sentence queries in open-world videos, there only exists very few works \cite{Hendricks2017Localizing,Gao2017TALL,liu2018attentive}, and all of them employ the aforementioned ``scan and localize' framework. Specifically, Hendricks \emph{at al.} presented a Moment Context Network (MCN) for matching candidate video clips and sentence query. In order to incorporate the contextual information, they enhanced the video clip representations by integrating both local and global video features overtime \cite{Hendricks2017Localizing}. To reduce the overload of scanning sliding windows, Gao \emph{at al.} proposed a Cross-modal Temporal Regression Localizer (CTRL) which only uses coarsely sampled clips, and then adjusts the locations of these clips by learning temporal boundary offsets through a temporal localization regression network. Inspired by the CTRL approach, Liu \emph{at al.} proposed a Attentive Cross-Modal Retrieval Network (ACRN) with two further extensions \cite{liu2018attentive}. The first is that they designed a memory attention mechanism to emphasize the visual feature mentioned in the query and incorporate it to the context of each samples clips. The second is that they enhanced the multi-modal representations of clip-query pairs with the outer product of their features. The proposed ABLR approach is fundamentally different from these three models, because they separately process each video clip from a local perspective, while our ABLR directly regresses the temporal coordinates of the target clip from a global view and can be trained in an end-to-end manner.

\begin{figure*}
\centering
\includegraphics[width=6.3in]{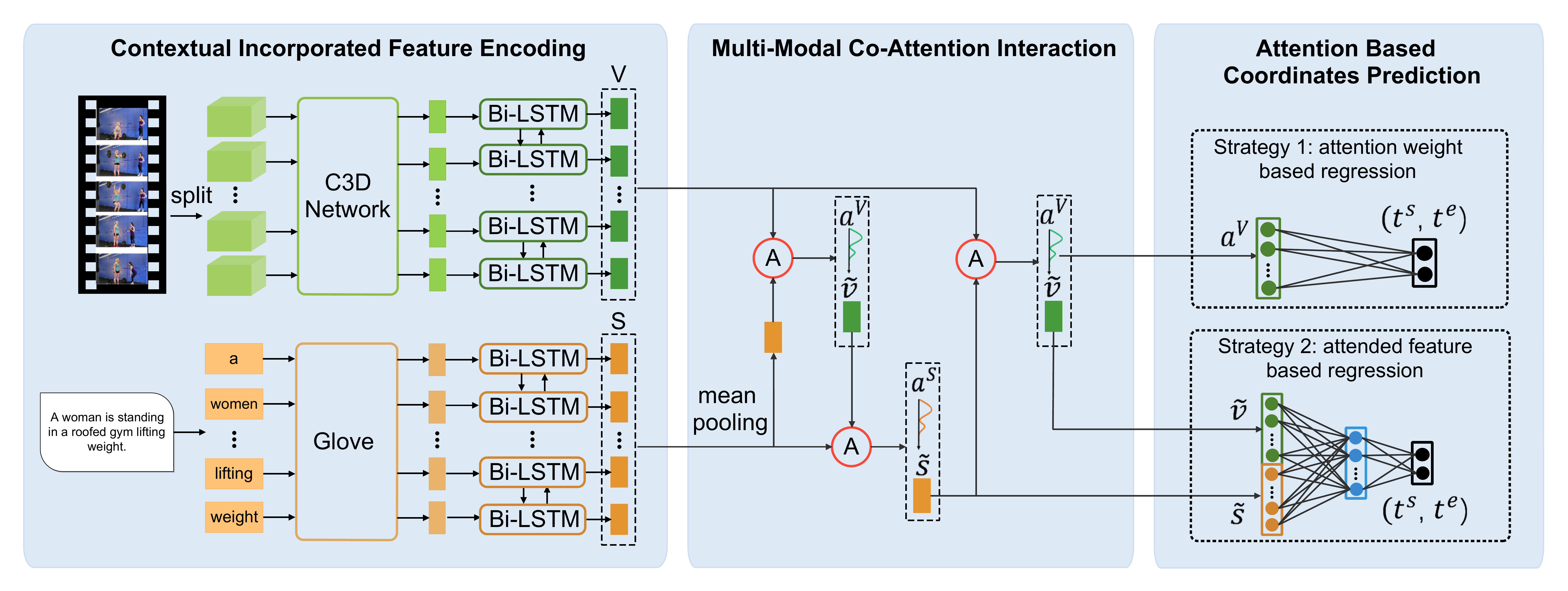}
\caption{ \small Framework of our end-to-end Attention Based Location Regression (ABLR) model. ABLR contains three main components. (1) Contextual Incorporated Feature Encoding preserves context information in video and sentence query through two Bi-directional LSTMs. (2) Multi-Modal Co-Attention Interaction sequentially alternates between generating video and sentence attentions. (3) Attention Based Coordinates Prediction sets up an attention based location prediction network, which regresses the temporal coordinates of sentence query from the former output attentions. Meanwhile, there are two different regression strategies: attention weight based regression and attended feature based regression.}
\label{fig:framwork}
\end{figure*}


\section{Attention Based Location Regression Model}
The main goal of our Attention Based Location Regression (ABLR) model is to build an efficient localization architecture which can locate the position of the sentence query in the overall video sequence. Meanwhile, the global video information and the concrete sentence details should be fully explored to ensure the localization accuracy. Therefore, as illustrated in Figure \ref{fig:framwork}, we design our ABLR model with three main components, i.e., contextual incorporated feature encoding which can preserve video and sentence contexts, multi-modal co-attention interaction which can generate global video attentions and highlight crucial sentence details, and attention based coordinates prediction which can directly regress the temporal coordinates of the target video segment. From the video and sentence inputs to the temporal coordinates output, the proposed ABLR model is an end-to-end architecture that can be jointly optimized.

In the following, we will first give the problem statement of the temporal sentence localization in video, and then present the three main components of ABLR in detail. Finally, we will introduce the learning procedure of the overall model.

\subsection{Problem Statement}
Suppose a video $V$ is associated with a set of temporal sentence annotations $\left\{ (S,\tau^{s},\tau^{e})\right\}$, where $S$ is a sentence description of a video segment with start and end time points $(\tau^{s}$,$\tau^{e})$ in the video. Given the input video and sentence, our task is to predict the corresponding temporal coordinates $ (\tau^{s}, \tau^{e})$.

\subsection{Contextual Incorporated Feature Encoding}
\textbf{Video Encoding}:
As temporal localization is to locate a position within the whole video, both specific video content and global video context are crucial elements that cannot be overlooked.
Some previous methods claim that they have incorporated contextual information in video clip features. However, they perform this through some hard-coding ways --- roughly fusing the global video features \cite{Hendricks2017Localizing} or extending the clip boundaries with a predefined scale \cite{Gao2017TALL,liu2018attentive}.
Actually, incorporating too much contextual information will confuse the localization procedure, and limiting the clip extension will fail to maintain some long-term relationships.
To overcome these problems, we propose to exploit Bi-directional LSTM network for video encoding.

For each untrimmed video $V$, we first evenly split it into M video clips $ \left\{ v_1, \cdots v_j, \cdots v_M \right\}$ in chronological order. Then, we apply the widely used C3D network \cite{Tran2014Learning} to encode these video clips. Finally, we use Bi-directional LSTM to generate video clip representations incorporated with the contextual information. Precise definitions are as follows:
\begin{equation} \label{video} \small
\begin{split}
\mathbf{x}_j &= C3D(v_j), \\
\mathbf{h}_j^f, \mathbf{c}_j^f &= LSTM^f(\mathbf{x}_j,\mathbf{h}^f_{j-1}, \mathbf{c}^f_{j-1}),\\
\mathbf{h}_j^b, \mathbf{c}_j^b &= LSTM^b(\mathbf{x}_j,\mathbf{h}^b_{j+1}, \mathbf{c}^b_{j+1}),\\
\mathbf{v}_j &= f \big(\mathbf{W}_v(\mathbf{h}_j^f \| \mathbf{h}_j^b)+\mathbf{b}_v \big).
\end{split}
\end{equation}

Here $\mathbf{x}_j$ is the fc7 layer C3D features of the video clip $v_j$. The Bi-directional LSTM consists of two independent streams, in which $LSTM^f$ moves from the start to the end of video and $LSTM^b$ moves end to start. The final representation $\mathbf{v}_j$ of video clip $v_j$ is computed by transforming the concatenation of the forward and backward LSTM outputs at position $j$. $f(\cdot)$ indicates the activation function, i.e., Rectified Linear Unit (ReLU), in this paper. As modeled in LSTM, adjacent video clips will influence each other, and therefore the representation of each video clip is enriched by a variably-sized context. The overall video can be represented as {\small $\mathbf{V} =  \left[ \mathbf{v}_1, \cdots \mathbf{v}_j, \cdots  \mathbf{v}_M \right] \in \mathbb{R}^{h_v \times M}$}, with each column indicating the final $h_v$-dimensional representation of a video clip.

\textbf{Sentence Encoding}:
In order to explore the details in sentence, we also employ Bi-directional LSTM to represent sentence as \cite{Karpathy2015Deep} did. Unlike the general LSTM which encodes sentence as a whole, the Bi-directional LSTM takes a sequence of N words $ \left\{ s_1, \cdots s_j, \cdots  s_N \right\}$ from sentence $S$ as inputs and encodes each word $s_j$ into a contextual incorporated feature vector $\mathbf{s}_j$. The precise definition of the sentence Bi-directional LSTM is similar to Eq (\ref{video}), and the input features are 300-D glove \cite{Pennington2014Glove} word features. Finally, the sentence representation is denoted as $ \mathbf{S} = \left[ \mathbf{s}_1, \cdots \mathbf{s}_j, \cdots  \mathbf{s}_N \right] \in \mathbb{R}^{h_s \times N}$. In the following, we unify the hidden state size of both sentence and video Bi-bidirectional LSTM as $h$, i.e., $h_v = h_s = h$.

\subsection{Multi-Modal Co-Attention Interaction}

In the literature, visual attention mechanism mainly focuses on the problem of identifying ``where to look'' on different visual tasks. As such, it can be naturally applied to temporal sentence localization, which is exactly to localize where to pay attention in video with the guidance of sentence description.
Furthermore, the problem of identifying ``which words to guide" or the sentence attention is also important for this task, as highlighting the key words or phrases will provide the localization procedure a more clear target.

Based on the above considerations, we propose to set up a symmetry interaction between video and sentence query by introducing the multi-modal co-attention mechanism \cite{Lu2016Hierarchical}. In this attention mechanism, we sequentially alternate between generating video and sentence attentions. Briefly, the process consists of three steps: (1) attend to the video based on the initial sentence feature; (2) attend to the sentence based on the attended video feature; (3) attend to the video again based on the attended sentence feature. Specifically, the attention function $\tilde{\mathbf{z}} = A(\mathbf{Z};\mathbf{g})$ takes the video (or sentence) feature $\mathbf{Z}$ and the attention guidance $\mathbf{g}$ derived from the sentence (or video) as inputs, and outputs the attended video (or sentence) feature as well as the attention weights. Concrete definitions are as follows:
\begin{equation} \label{attention} \small
\begin{split}
\mathbf{H} &=  tanh \big(\mathbf{U}_z\mathbf{Z}+(\mathbf{U}_g\mathbf{g})\mathbf{1}^T+\mathbf{b}_a\mathbf{1}^T \big), \\
\mathbf{a}^z &= softmax(\mathbf{u}_a^T\mathbf{H}), \\
\tilde{\mathbf{z}} &= \sum a_j^z \mathbf{z}_j.
\end{split}
\end{equation}

Here $\mathbf{U}_g, \mathbf{U}_z \in \mathbb{R}^{k \times h}$, $\mathbf{b}_a, \mathbf{u}_a \in \mathbb{R}^{k}$ are parameters of the attention function, $\mathbf{1}$ is a vector with all elements to be 1, $\mathbf{a}^z$ is the attention weights of $\mathbf{Z}$, $\tilde{\mathbf{z}}$ is the attended feature. As shown in the middle part of Figure \ref{fig:framwork}, in the first step of alternative attention, $\mathbf{Z} = \mathbf{V}$ and $\mathbf{g}$ is the average representation of words in the sentence. In the second step, $\mathbf{Z} = \mathbf{S}$ and $\mathbf{g}$ is the intermediate attended video feature from the first step. In the last step, we attend the video again based on the attended sentence feature from the second step.

Through the above process, the video attention weights $\mathbf{a}^V$ can be considered as a kind of feature in temporal dimension, in which one single element $a^V_j$ represents the relative association between the $j$th video clip and the sentence description. Therefore, the entire video attention weights will reflect the global temporal structure of the video and the attended video feature will focus more on the specific video contents which are relevant to the sentence description. Meanwhile, since we also calculate the sentence attention based on the video content, the crucial words and phrases in the sentence will provide a stronger guidance in the localization procedure.

\subsection{Attention Based Coordinates Prediction}
Given the video attention weights $\mathbf{a}^V$ produced from the last step of co-attention function, one possible way to localize the sentence is to choose or merge some video clips with higher attention values \cite{Rohrbach2016Grounding,Zheng2017CDC}. However, these methods rely on some post-processing strategies and separate the location prediction with the former modules, resulting in sub-optimal solutions. To avoid this problem, we propose a novel attention based location prediction network, which directly explores the correlation between the former attention outputs and the target location. Specifically, the attention based location prediction network takes the video attention weights or the attended features as input, and regresses the normalized temporal coordinates of the selected video segment. In addition, we design two kinds of location regression strategies: one is attention weight based regression and the other is attended feature based regression.

Attention weight based regression takes the video attention weights $\mathbf{a}^{V}$ as a kind feature in temporal dimension and directly regresses the normalized temporal coordinates~as:
\begin{equation} \label{regession_2} \small
\mathbf{t} =({t}^{s},{t}^{e}) = f \big(\mathbf{W}_{aw} (\mathbf{a}^{V})^T+\mathbf{b}_{aw} \big).
\end{equation}
Here $\mathbf{W}_{aw} \in \mathbb{R}^{2 \times M}$ and $\mathbf{b}_{aw} \in \mathbb{R}^2$ are regression parameters. $({t}^{s},{t}^{e})$ are the predicted start and end times of the sentence description, which points out the position in video.

Attended feature based regression firstly fuses the attended video feature $\tilde{\mathbf{v}}$ and sentence feature $\tilde{\mathbf{s}}$ to a multi-modal representation $\mathbf{f}$, and then regresses the temporal coordinates as:
\begin{equation} \label{regession_3} \small
\begin{split}
 \mathbf{f} &= f \big(\mathbf{W}_f(\tilde{\mathbf{v}} \| \tilde{\mathbf{s}})+\mathbf{b}_f \big),\\
\mathbf{t} &= ({t}^{s},{t}^{e}) = f(\mathbf{W}_{af}\mathbf{f}+\mathbf{b}_{af}).
\end{split}
\end{equation}
Here $\mathbf{W}_{f} \in \mathbb{R}^{h \times 2h}$ and $\mathbf{b}_{f} \in \mathbb{R}^{h}$ are used for feature fusion. $\mathbf{W}_{af} \in \mathbb{R}^{2 \times h}$ and $\mathbf{b}_{af} \in \mathbb{R}^2$ are parameters of the attended feature based regression. As shown in Eq (\ref{regession_2}) and Eq (\ref{regession_3}), we define the temporal coordinates regression with the form of a single layer fully connected operation. Practically, there are two fully connected layers between the input data and the output temporal coordinates in our settings.

Both the two regression strategies consider the entire video environment as the temporal localization basis, but are suitable for different video scenarios. We will discuss the influence of them in the Experiments section.

\subsection{Learning of ABLR}
Firstly, we denote the training set of ABLR as $\left\{ (V_i,\tau_i,S_i,\tau_i^s, \tau_i^e) \right\}_{i=1}^K$. $V_i$ is a video of duration $\tau_i$. $S_i$ is a sentence description of a particular video segment, which has start and end points $(\tau_i^s,\tau_i^e)$ in video $V_i$. Note that one video can have multiple sentence descriptions, and therefore different training samples may refer to the same video.

We normalize the start and end time points of sentence $S_i$ to $\tilde{\mathbf{t}_i} = (\tilde{t}^{s}_i, \tilde{t}^{e}_i) =(\tau^{s}_i / \tau_i, \tau^{e}_i / \tau_i)$, which is regarded as the ground truth of the location regression.
With the ground truth and predicted temporal coordinates pair $(\mathbf{\tilde{t}}_i,\mathbf{t}_i)$, we design an \textbf{attention regression loss} to optimize the temporal coordinates prediction, which is defined by the form of smooth L1 function R(x) \cite{Girshick2015Fast}:
\begin{equation} \label{regession_4} \small
L_{reg} =  \sum_{i=1}^K [R(\tilde{t}^{s}_i-t^{s}_i)+R(\tilde{t}^{e}_i-t^{e}_i)].
\end{equation}

In general, the learning procedure of attention does not have any explicit ground truth of attention to guide, no matter in \cite{Lu2016Hierarchical} or other computer vision tasks. However, in ABLR, we need to directly regress the temporal coordinates from video attentions. Therefore, the accuracy of the learned video attentions will have a great influence on the subsequent location regression. Based on the above consideration, we additionally design an \textbf{attention calibration loss}, which constrains the multi-modal co-attention module to generate video attentions well aligned with the ground truth temporal interval:
\begin{equation} \label{regession_5} \small
L_{cal} = - \sum_{i=1}^K \frac{\sum_{j=1}^M m_{i,j} log(a_j^{{V}_i})}{\sum_{j=1}^M m_{i,j}}.
\end{equation}
Here $m_{i,j} = 1$ indicates that the $j$th video clip in video $V_i$ is within the ground truth window $(\tau_i^s,\tau_i^e)$ of sentence $S_i$, otherwise $m_{i,j} = 0$. Obviously, the attention calibration loss encourages the video clips within ground truth windows to have higher attention values. For sentence attention, as there is lack of annotations, we can only implicitly learn it from the overall model as in general cases.

The overall loss of our localization system consists of both the attention regression and the attention calibration loss:
\begin{equation} \label{regession_6} \small
L = \alpha L_{reg} + \beta L_{cal}.
\end{equation}
$\alpha$ and $\beta$  are hyper parameters which control the weights between the two loss terms, and the values of them are determined by grid search.

With the above overall loss term, our ABLR model can be trained end to end from the feature encoding step to the coordinates prediction step. In test stage, we input the video and sentence query to our ABLR model and then output the normalized temporal coordinates of the sentence query. During this process, we obtain the absolute position by multiplying the normalized temporal coordinates with video duration. Since the video attention weights are calculated at clip level, we finally trim the predicted interval to include integer numbers of video clips.

\section{Experiments}

\subsection{Datasets}

In this work, two public datasets are exploited for evaluation.

\textbf{TACoS} \cite{Regneri2013Grounding}: Textually Annotated Cooking Scenes (TACoS) contains a set of video descriptions (in natural language) and timestamp-based alignment with the videos. In total, there are 127 videos  picturing people who perform cooking tasks, and approximately 17000 pairs of sentences and video clips. We use 50\% of the dataset for training, 25\% for validation and 25\% for test as \cite{Gao2017TALL} did.

\textbf{ActivityNet Captions} \cite{Krishna2017Dense}: This dataset contains 20k videos with 100k descriptions, each with a unique start and end time. Compared to TACoS, ActivityNet Captions has two orders of magnitude more videos and provides annotations for an open domain. The public training set is used for training, and validation set for testing.

\subsection{Experimental Settings}

\textbf{Compared Methods}: Since temporal sentence localization in video is a new research direction, there are few existing works to compare with and we list them as follows:

(1) MCN \cite{Hendricks2017Localizing}:  Moment Contextual Network as mentioned before.

(2) CTRL \cite{Gao2017TALL}: Cross-modal Temporal Regression Localizer as mentioned before.

(3) ACRN \cite{liu2018attentive}: Attentive Cross-Modal Retrieval Network as mentioned before.

To validate the effectiveness of our ABLR design, we also ablate our model with different configurations as follows.

(1) ABLP: Attention Based Localization by Post-processing the video attentions. Attention based location regression strategies are omitted in this variant. Temporal coordinates of sentence queries are determined by applying the temporal boundary refinement \cite{Zheng2017CDC} on video attention weights.

(2) ABLR$_{\textbf{reg-aw}}$/ABLR$_{\textbf{reg-af}}$: ABLR model which is trained with attention regression loss only, attention calibration loss is omitted. In addition, ``aw'' means attention weight based regression strategy is adopted and ``af'' means attended feature based regression is adopted.

(3) ABLR$_{\textbf{c3d-aw}}$/ABLR$_{\textbf{c3d-af}}$: We remove the Bi-directional LSTM in video encoding. Video clips are represented by C3D features, without incorporating the contextual information.

(4) ABLR$_{\textbf{stv-aw}}$/ABLR$_{\textbf{stv-af}}$: The Bi-directional LSTM for sentence encoding is replaced by Skip-thought \cite{Kiros2015Skip} sentence embedding extractor. Therefore, each sentence description is represented by a single feature vector, and the proposed multi-modal co-attention module is degraded with only video attention reserved.

(5) ABLR$_{\textbf{full-aw}}$/ABLR$_{\textbf{full-af}}$: Our full ABLR model.

\textbf{Evaluation Metrics}: We adopt similar metrics ``R@1, IoU@$\sigma$" and ``mIoU" from \cite{Gao2017TALL} to evaluate the performance of temporal sentence localization. For each sentence query, we calcuate the Intersection over Union (IoU) between the predicted and ground truth temporal coordinates. ``R@1, IoU@$\sigma$" means the percentage of the sentence queries which have IoU larger than $\sigma$. Meanwhile, ``mIoU" means the average IoU for all the sentence queries.

\textbf{Implementation details}: Based on the video duration distribution, videos in ActivityNet Captions and TACoS dataset are averagely split into 128 clips and 256 clips, respectively. Additionally, we choose the hidden state size $h$ of both video and sentence Bi-directional LSTM as 256, dropout rate as 0.5 through ablation studies. As for multi-modal co-attention module, we set the hidden state size $k$ = 256, therefore $\mathbf{U}_g, \mathbf{U}_z \in \mathbb{R}^{256 \times 256}$, $\mathbf{u}_a, \mathbf{b}_a \in \mathbb{R}^{256}$. The trade-off parameters $\alpha$ and $\beta$ are set as 1 and 5 by grid search. We train our model using a mini-batch of 100 and learning rate of 0.001 for ActivityNet Captions, 0.0001 for TACoS.

\begin{table}[!tb]\small
\centering
\caption{ \small Comparison of different methods on ActivityNet Captions}
\begin{tabular}{m{1.9cm} m{1cm}<{\centering} m{1cm}<{\centering} m{1cm}<{\centering} m{1cm}<{\centering}}
\hline
Methods & R@1, IoU@0.1 & R@1, IoU@0.3  & R@1, IoU@0.5  & mIoU \\
\hline
MCN  &  0.4280 & 0.2137 & 0.0958 & 0.1583\\
CTRL & 0.4909 & 0.2870 & 0.1400 & 0.2054  \\
ACRN & 0.5037 & 0.3129 & 0.1617 & 0.2416 \\
\hline
ABLP             & 0.6804 & 0.4503  & 0.2304 & 0.2917 \\
ABLR$_{reg-af}$  & 0.6922 & 0.5184  & 0.3353 & 0.3509 \\
ABLR$_{reg-aw}$  & 0.6982 & 0.5187  & 0.3339 & 0.3501 \\
ABLR$_{c3d-af}$  & 0.6810 & 0.5113  & 0.3161 & 0.3356 \\
ABLR$_{c3d-aw}$  & 0.5578 & 0.3943  & 0.1835 & 0.2434 \\
ABLR$_{stv-af}$  & 0.6841 & 0.5083  & 0.3153 & 0.3368 \\
ABLR$_{stv-aw}$  & 0.7177 & 0.5442  & 0.3094 & 0.3426 \\
ABLR$_{full-af}$ & 0.7023 & 0.5365  & 0.3491 & 0.3572 \\
ABLR$_{full-aw}$ & \textbf{0.7330} & \textbf{0.5567}  & \textbf{0.3679} & \textbf{0.3699} \\
\hline
\end{tabular}
\label{variants}
\end{table}

\subsection{Evaluation on ActivityNet Captions Dataset}
Table \ref{variants} shows the R@1,IoU@$\{0.1,0.3,0.5\}$ and mIoU performance of different methods on ActivityNet Captions dataset. Overall, the results consistently indicate that our ABLR model outperforms others. Notably, the mIoU of ABLR$_{full-aw}$ makes a significant improvement over the best competitor ACRN relatively by 53.1$\%$, which validates the effectiveness of our ABLR design. Among all the baseline methods, we could see that MCN receives worse results compared to others. We speculate the main reason is that MCN treats the average pooling of all the video clip features as the context representation of each candidate clip, ignoring the relative importance of different video contents. Roughly fusing global video context with local video features like MCN will bring some noisy information to the localization procedure, and therefore influence the localization accuracy. Both CTRL and ACRN derive from the typical ``scan and localize'' architecture, i.e., firstly splits a video clip candidate from the whole video and then adjusts the clip boundary to the target position. Although these two approaches enhance the video clip features with some predefined neighboring context, they overlook the global temporal structure and possible long-term relationships within videos, and therefore do not achieve satisfying results.

\begin{figure}[!tb]
\centering
\includegraphics[width=3.4in]{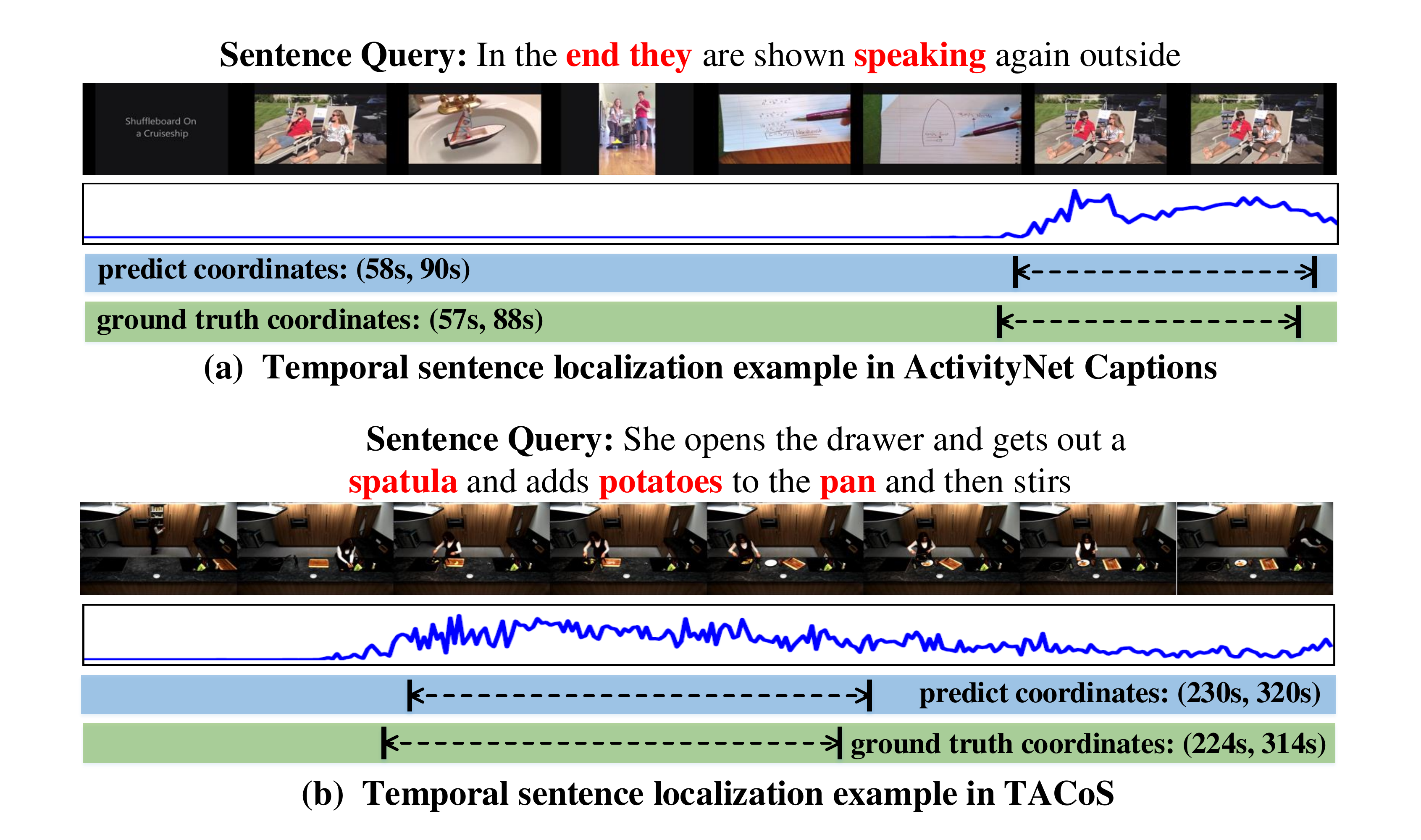}
\caption{\small Qualitative results of ABLR model for temporal sentence localization. The bar with blue background shows the predicted temporal interval for the sentence query, the bar with green background shows the ground truth. Video attention weights are represented with the blue waves in the second bar, words of high attention weights in sentence queries are highlighted by red font.}
\label{fig:qualitative}
\end{figure}

To better demonstrate the superiority of ABLR, we provide some qualitative results in Figure \ref{fig:qualitative}. Specifically, as shown in Figure \ref{fig:qualitative}(a), the scene describing the people speaking behavior appears twice in the video, while the sentence query provides an important cue ``in the end'', and therefore the latter one is the correct location. However, it is hard for the previous methods to make a good decision, because they process each candidate clip independently and do not explore the relative relations between the local video part and the global video environment.  In contrast, our ABLR model not only maintains the adaptive contextual information through Bi-directional LSTM, but also preserves the global temporal structure of video through the video attention outputs. Therefore, the localization decision making in ABLR is more accurate and comprehensive. Furthermore, with the multi-modal co-attention mechanism, we can also see from Figure \ref{fig:qualitative} that the learned sentence attentions highlight some key words in sentence queries, such as some objects, actions and even words with time meaning. These highlighted words provide clear cues for localizing, and enhance the interpretability of the localization system.

As for different configurations of our ABLR model, we could see from Table \ref{variants} that in both attention weight based regression and attended feature based regression strategies, our full ABLR design ABLR$_{full}$ substantially outperforms other variants. In particular, the mIoU of ABLR$_{full-af}$ makes the relative improvement over ABLR$_{c3d-af}$ by 6.4$\%$, which proves the importance of video contextual information in temporal localization. Comparing ABLR$_{stv-aw}$ with ABLR$_{full-aw}$, we can see the introduction of sentence attention brings up to 18.9$\%$ improvement in terms of R@1,IoU@0.5. It shows that considering the sentence details and emphasizing the key words in sentence queries can increase the localization accuracy. By incorporating attention calibration loss, ABLR$_{full}$ exhibits better performance than ABLR$_{reg}$. It demonstrates the advantage of attention calibration, which encourages the multi-modal co-attention module to learn video attentions well aligned with the temporal coordinates, and further provides more accurate inputs for the location regression network. Moreover, there is a performance degradation from ABLR$_{full}$ to ABLP. The results confirm the superiority of the attention based location regression over the post processing strategy. Also noted that the experimental results of different configurations in TACoS are consistent with ActivityNet Captions.

\begin{figure}[!tb]
\centering
\includegraphics[height=2.3in]{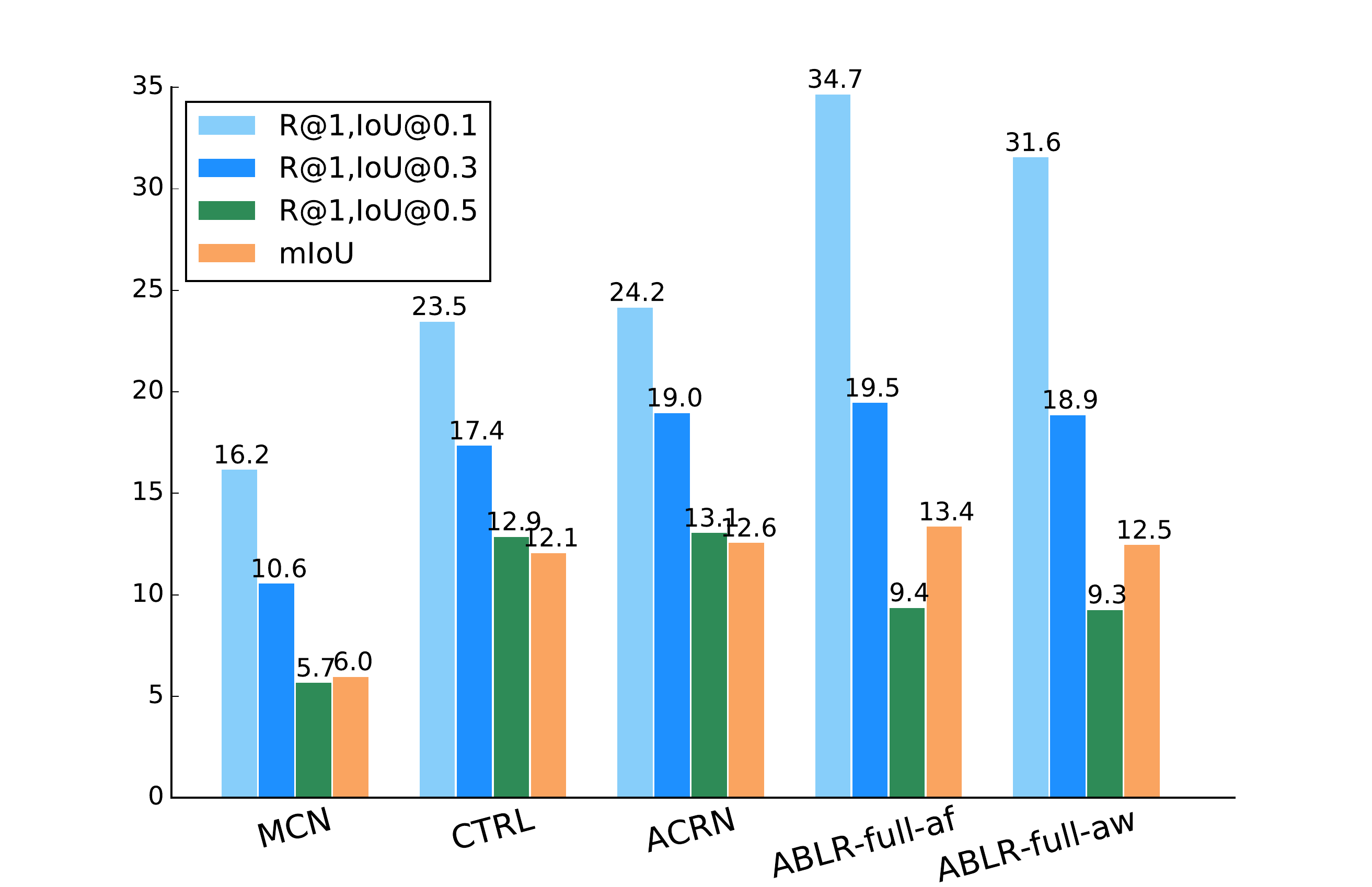}
\caption{\small Comparison of different methods on TACoS}
\label{fig:compare_tacos}
\end{figure}

\subsection{Evaluation on TACoS Dataset}

We also test our ABLR model and baseline methods on TACoS dataset, and report the results in Figure \ref{fig:compare_tacos}. Overall, it can be observed that the results are consistent with those in ActivityNet Captions, i.e., ABLR outperforms other baseline methods. Meanwhile, we can also see some other interesting observations.

The first observation is that the R@1,IoU@0.1 and R@1,IoU@0.3 performance of our ABLR$_{full-af}$ makes the relative improvement over the best competitor ACRN by 43.4$\%$ and 2.6$\%$ respectively, while R@1 is below that of ACRN when the IoU threshold increases to 0.5.
We speculate this phenomenon is caused by the obvious distinction between the two datasets. As shown in Figure \ref{fig:qualitative}, videos in ActivityNet Captions contain various scenes and activities, and even in a single video, the variance between different segments is obvious. Instead, all the videos in TACoS share a common scene, and only the people and the cooking objects are changed. Indistinguishable video clips in TACoS result in a relatively flatter attention wave. Under this condition, the ABLR model can effectively locate the approximate position of the sentence query, achieving better results of R@1 at lower IoU value, but will be confused to determine the precise segment boundaries with the requirement of higher IoU threshold. As for CTRL and ACRN, they split candidate video clips from the whole video and compare the sentence query with each of these clips individually. Therefore, they can reduce the disturbance caused from similar scenes in TACoS videos. However, the splitting strategy of CTRL and ACRN makes their localization efficiency much lower than ABLR, which will be further discussed in the Efficiency Analysis section. Moreover, compared with ActivityNet Captions dataset which have 20k videos, the TACoS dataset have only 127 videos. The limited size of TACoS restricts the training procedure of the localization methods, and will also influence the model performance.

The second observation is that attention weight based regression ABLR$_{full-aw}$ outperforms attended feature based regression ABLR$_{full-af}$ on ActivityNet Captions, while ABLR$_{full-af}$ is better on TACoS. Since ABLR$_{aw}$ directly regresses temporal coordinates from video attention weights, the flat and ambiguous attention waves in TACoS make ABLR$_{aw}$ hard to determine sentence positions accurately. Compared to ABLR$_{aw}$, ABLR$_{af}$ incorporates video content information and further enhances the discriminability of the inputs of the location regression network, and thus leads to better results.

\subsection{Efficiency Analysis}
Table \ref{table:compare_time} shows the average run time to localize one sentence in video for different methods. Compared with MCN, CTRL and ACRN, our ABLR significantly reduces the localization time by a factor of 4 $\sim$ 15 in ActivityNet Captions dataset, and the advantage is more obvious when localizing sentence query in longer videos from TACoS dataset. The results verify the merit of the proposed ABLR model. Previous methods which adopt the two stage ``scan and localize'' architecture, often need to sample densely overlapped video clip candidates by various sliding windows. Therefore, they have to process a large number clips one by one to localize sentence queries. However, ABLR model only needs to pass through each video twice in the video encoding procedure, and thus it can avoid redundant computations. All the experiments are conducted on an Ubuntu 16.04 server with Intel Xeon CPU E5-2650, 128 GB Memory and NVidia Tesla M40 GPU.

\begin{table}[!tb] \small
\centering
\caption{ \small Average time to localize one sentence for different methods}
\begin{tabular}{m{1.7cm}<{\centering} | m{1.7cm}<{\centering} m{1.7cm}<{\centering} }
\hline
 Methods & ActivityNet Captions & TACoS    \\
\hline
MCN  & 0.30s & 9.41s  \\
CTRL & 0.09s & 3.75s  \\
ACRN & 0.12s & 5.29s \\
ABLR &  \textbf{0.02s} & \textbf{0.15s} \\
\hline
\end{tabular}
\label{table:compare_time}
\end{table}

\section{Conclusions and Future Work}
In this paper, we address the problem of temporal sentence localization in untrimmed videos, and propose a Attention Based Location Regression (ABLR) model, which is a novel end-to-end architecture for solving the localization problem. The ABLR model with multi-modal co-attention mechanism not only learns the video attentions reflecting the global temporal structure, but also explores the crucial sentence details for localization. Furthermore, the proposed attention based location prediction network directly regresses the temporal coordinates from the global attention outputs, which avoids the trivial post processing strategy and makes the overall model can be globally optimized.
With all the above designs, our ABLR model is able to achieve superior localization accuracy on both ActivityNet Captions and TACoS dataset, and significantly boosts the localization efficiency.

Further problems, like temporal localization of multiple sentences in videos and sentence localization in both temporal and spatial dimensions, can also be investigated in the future.

\section{Acknowledgements}
This work was supported in part by National Program on Key Basic Research Project (No.2015CB352300), and National Natural Science Foundation of China Major Project (No. U1611461). The authors would like to thank Dr. Ting Yao, Dr. Jun Xu, Linjun Zhou and Xumin Chen for their great supports and valuable suggestions on this work.


\end{document}